\documentclass[letterpaper, 10 pt, conference]{ieeeconf}  

\IEEEoverridecommandlockouts                              

\usepackage{graphicx}
\usepackage{subcaption}
\usepackage{amsmath}

\title{\LARGE \bf
Biomimetic Mantaray robot toward the underwater autonomous - Experimental verification of swimming and diving by flapping motion -
}

\author{Kenta Tabata$^{1}$, Ryosuke Oku$^{1}$, Jun Ito$^{1}$, Renato Miyagusuku$^{1}$ and Koichi Ozaki$^{1}$
\thanks{*The research was financially supported by the Sasakawa Scientific Research
 Grant from The Japan Science Society.}
\thanks{$^{1}$Kenta Tabata, Ryosuke Oku, Jun Ito, Renato Miyagusku and Koichi Ozaki are with Department of Mechanical and Intelligent Engineering, 
        Utsunomiya University, Utsunomiya, Tochigi, Japan
        {\tt\small tabata@cc.utsunomiya-u.ac.jp}}%
}

\begin{document}

\maketitle
\thispagestyle{empty}
\pagestyle{empty}

\begin{abstract}

This study presents the development and experimental verification of a biomimetic manta ray robot for underwater autonomous exploration. Inspired by manta rays, the robot uses flapping motion for propulsion to minimize seabed disturbance and enhance efficiency compared to traditional screw propulsion.
The robot features pectoral fins driven by servo motors and a streamlined control box to reduce fluid resistance. The control system, powered by a Raspberry Pi 3B, includes an IMU and pressure sensor for real-time monitoring and control.
Experiments in a pool assessed the robot's swimming and diving capabilities. Results show stable swimming and diving motions with PD control. The robot is suitable for applications in environments like aquariums and fish nurseries, requiring minimal disturbance and efficient maneuverability.
Our findings demonstrate the potential of bio-inspired robotic designs to improve ecological monitoring and underwater exploration.

\end{abstract}

\section{Introduction}

Underwater robots have been developed for exploring underwater environments. Many types of underwater robots are equipped with screw propulsion mechanisms due to their high propulsion efficiency and ease of control underwater. However, screw mechanisms generate turbidity by stirring up muddy sand from the sea bottom. This makes screw mechanisms unsuitable for underwater exploration and monitoring using cameras. A bio-inspired robot is a promising approach to solving this problem. Aquatic organisms (e.g., manta rays, turtles, jellyfish, octopuses, etc.) use fins and tentacles for propulsion. Generally, the motion frequency of aquatic organisms is slower than that of screw propellers. Therefore, they can move without stirring up muddy sand, and their motion is known for its high efficiency\cite{Habib2013EngineeringCD}.

We developed an underwater robot inspired by the manta ray. The main target application is environments with gradual flow, such as aquariums or fish nurseries. In this paper, we report on the hardware configuration, swimming algorithm, and the results of diving and swimming motion experiments.
\section{Related Works}
\subsection{Classification of bio-inspired under water robot}

Currently, various bio-mimetic underwater robots inspired by different aquatic organisms are being developed \cite{salazar2018classification}. 
Aquatic organisms exhibit a wide range of swimming methods and body shapes. When applying these characteristics to robots, the features can vary significantly depending on the organism being mimicked. 
C.M. Breder classified fish swimming methods and tail fin shapes into three categories: mackerel type, eel type, and boxfish type \cite{bhlpart203769}. 
The mackerel type swims by oscillating the tail fin and the posterior part of the body, achieving high speeds, as seen in tuna and bonito. 
The eel type swims by undulating its elongated body. The boxfish type swims by primarily moving its tail fin and pectoral fins, with minimal body movement.
Later, Lindsey further refined the classification of swimming methods into two categories: BCF (Body and/or Caudal Fin locomotion) propulsion and MPF (Median and/or Pectoral Fin locomotion) propulsion \cite{LINDSEY19781}. 
BCF propulsion, used by mackerel and eel types, involves swimming with the tail fin and body, and is advantageous for high-speed swimming. 
MPF propulsion, used by the boxfish type, primarily involves the tail fin, pectoral fins, and dorsal fin, and excels in roll and pitch control.

Underwater robots must operate under various disturbances such as currents and waves during ecological surveys and sea bottom exploration, and they must also avoid obstacles like rocks and floating debris. 
Additionally, when observing aquatic organisms, it is crucial to have good posture control to follow the organisms. Therefore, we focused on MPF (Median and/or Pectoral Fin) propulsion, which excels in posture control.
Among these, the manta ray can generate significant propulsion by slowly flapping its large pectoral fins. It can also perform maneuvers such as somersaults and jumps out of the water. Additionally, the manta ray can glide like a glider by stopping the flapping motion of its pectoral fins, allowing for energy-efficient and highly versatile swimming . In this study, we considered the manta ray to be the most suitable model for our underwater robot due to these characteristics.

\subsection{Previous developed manta ray robots}

Research on the hardware development of manta ray robots employs diverse approaches. The most common focus is on the design and material selection to mimic the morphology and motion characteristics of manta rays. 
Claudio  designed a flexible manta ray robot using 3D printing technology and tested its initial prototype in both air and water \cite{claudio2020design}. 
This prototype's performance was evaluated to collect data for further improvements. 
Similarly, Wang et al. proposed a soft manta robot using bidirectional dielectric elastomer muscle actuators, achieving movements close to the natural 3D morphology of manta rays \cite{wang2024soft}. 
Zhou et al. developed a biomimetic underwater robot called RoMan-II, which introduced a design utilizing flexible fins to generate propulsion \cite{zhou2010better}. 
This robot incorporated bio-inspired gliding motions to enhance energy efficiency, resulting in more efficient movement. Additionally, Asada et al. developed a manta ray robot with underwater walking capabilities, designing the pectoral fins into six leg fins by dividing each pectoral fin into three parts \cite{asada2024development}. 
This design enables efficient movement with low energy consumption without increasing system complexity. 
Furthermore, Osorio et al. proposed a multi-stable soft robot, developing a structure with multiple stable states actuated by dome-shaped units operated by pneumatic pressure. This design achieved fin deformation and stable shapes \cite{osorio2023manta}.

Research on the motion control algorithms of manta ray robots is also diverse, exploring methods to bring the robot's movement patterns closer to those of natural manta rays. 
Cao et al. achieved gliding and flapping propulsion in a manta robot using a CPG-based biomimetic motion control method and a PSO-based online similarity optimization method \cite{cao2023realization}. 
This approach enhanced the motion performance of the manta robot. 
Additionally, Liu et al. employed an improved YOLOv5 object detection algorithm and binocular stereo matching algorithm for the recognition and localization of sea cucumbers \cite{liu2023design}. 
This method enabled high-precision visual recognition and localization of sea cucumbers. 
Furthermore, Zhou et al. ~\cite{zhou2010better} adopted non-harmonic gait control to mimic the natural movements of manta rays, aiming to reproduce more realistic motion patterns compared to traditional harmonic control.

Research on the social implementation of manta ray robots explores methods to expand their potential applications. 
As indicated by most studies, manta robots have significant potential in ocean exploration and environmental monitoring. 
For example, Liu et al. ~\cite{liu2023design} demonstrated that manta robots could be useful for real-time monitoring and automated harvesting in sea cucumber aquaculture. 
Many studies suggest that manta ray robots could also be applied in military contexts, including covert underwater reconnaissance and surveillance operations. 
Additionally, research by Claudio \cite{claudio2020design} and Wang et al. \cite{wang2024soft} indicates that mimicking the motion characteristics of manta rays could inspire the design of other underwater robots and flexible electronic devices. 
Furthermore, Zhou et al. \cite{zhou2010better} showed that by mimicking the motion characteristics of manta rays, the hydrodynamic performance of these robots was improved compared to conventional ROVs.
These studies demonstrate diverse approaches to the design and control of manta ray robots, providing a foundation for expanding their applications. 
By combining flexible materials with multi-stable structures and developing advanced control algorithms, manta ray robots are expected to find applications in various fields.

\section{Developed manta ray robot}
\begin{figure}[h]
 \centering
  \includegraphics[height=0.5\hsize]{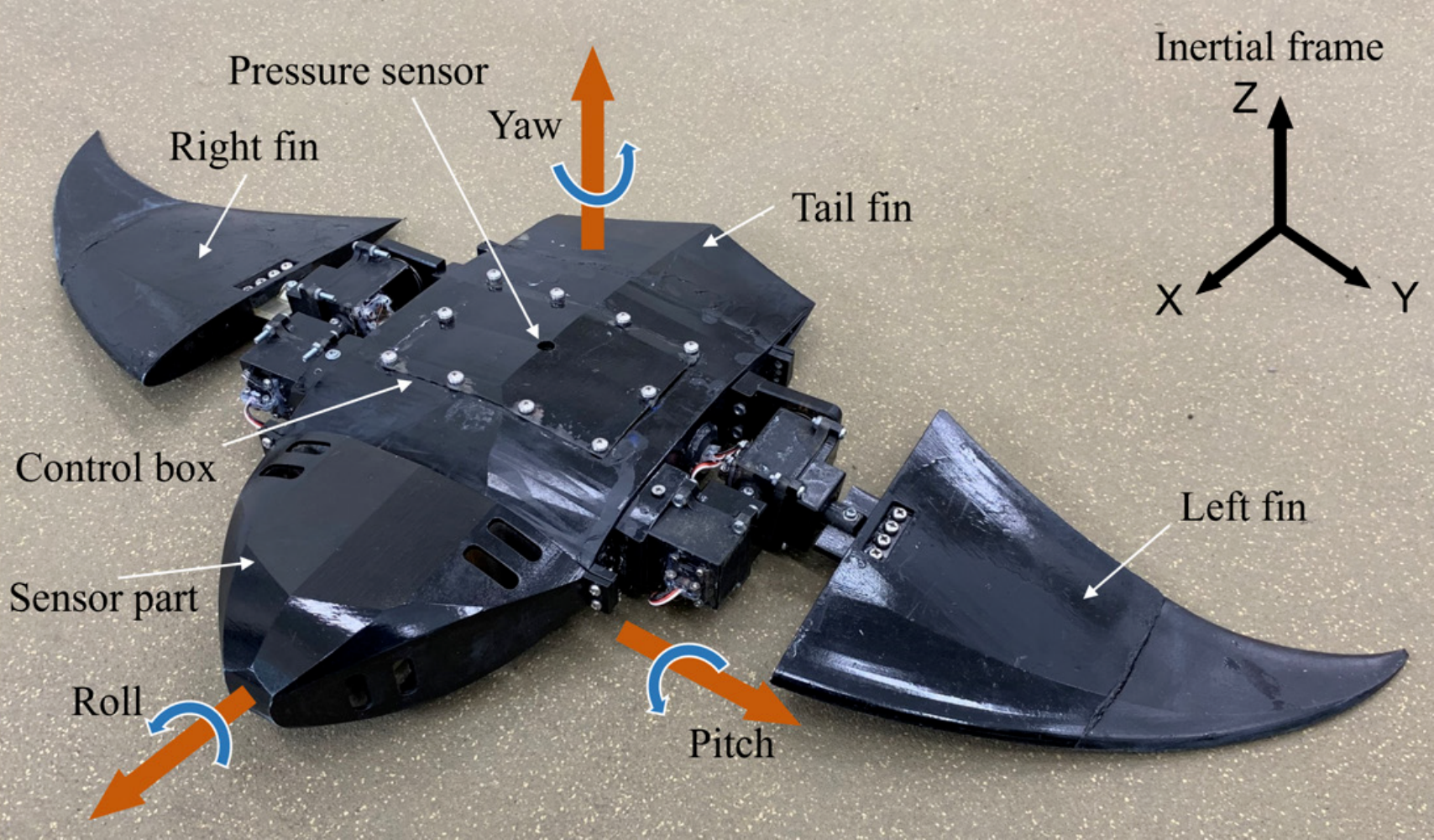}
  \caption{Developed manta ray robot}
  \label{fig:DevelManta}
\end{figure}

The developed robot is shown in {\bf{\ref{fig:DevelManta}}}, with its dimensions and weight listed in {\bf{Table \ref{tb:DevelMantaConf}}}. 
The robot consists of four modules: Sensor part, Control box, Right fin, Left fin, and Tail fin. 
Each pectoral fin is equipped with two servo motors (RS303MR by Futaba Corporation), allowing for a flapping motion around the X-axis and a feathering motion around the Y-axis, as illustrated in {\bf{\ref{fig:DevelManta}}}. This combination of movements simulates the manta ray's flapping motion. By adjusting the amplitude and phase of the flapping, the robot can achieve both straight and turning swimming motions.

\begin{table}[h!]
    \centering
    \caption{Manta ray robot dimensions and weight}
    \scalebox{0.9}{
    \begin{tabular}{|c|c|c|c|}
    \hline
    Length [mm] & Width [mm] & Height [mm] & Weight [g] \\ \hline
    360         & 750        & 70          & 2150       \\ \hline
    \end{tabular}
    }
    \label{tb:DevelMantaConf}
\end{table}

\subsection{Hardware configuration}

\begin{figure}[h]
 \centering
    \begin{minipage}[h]{\hsize}
        \centering
      \includegraphics[height=0.5\hsize]{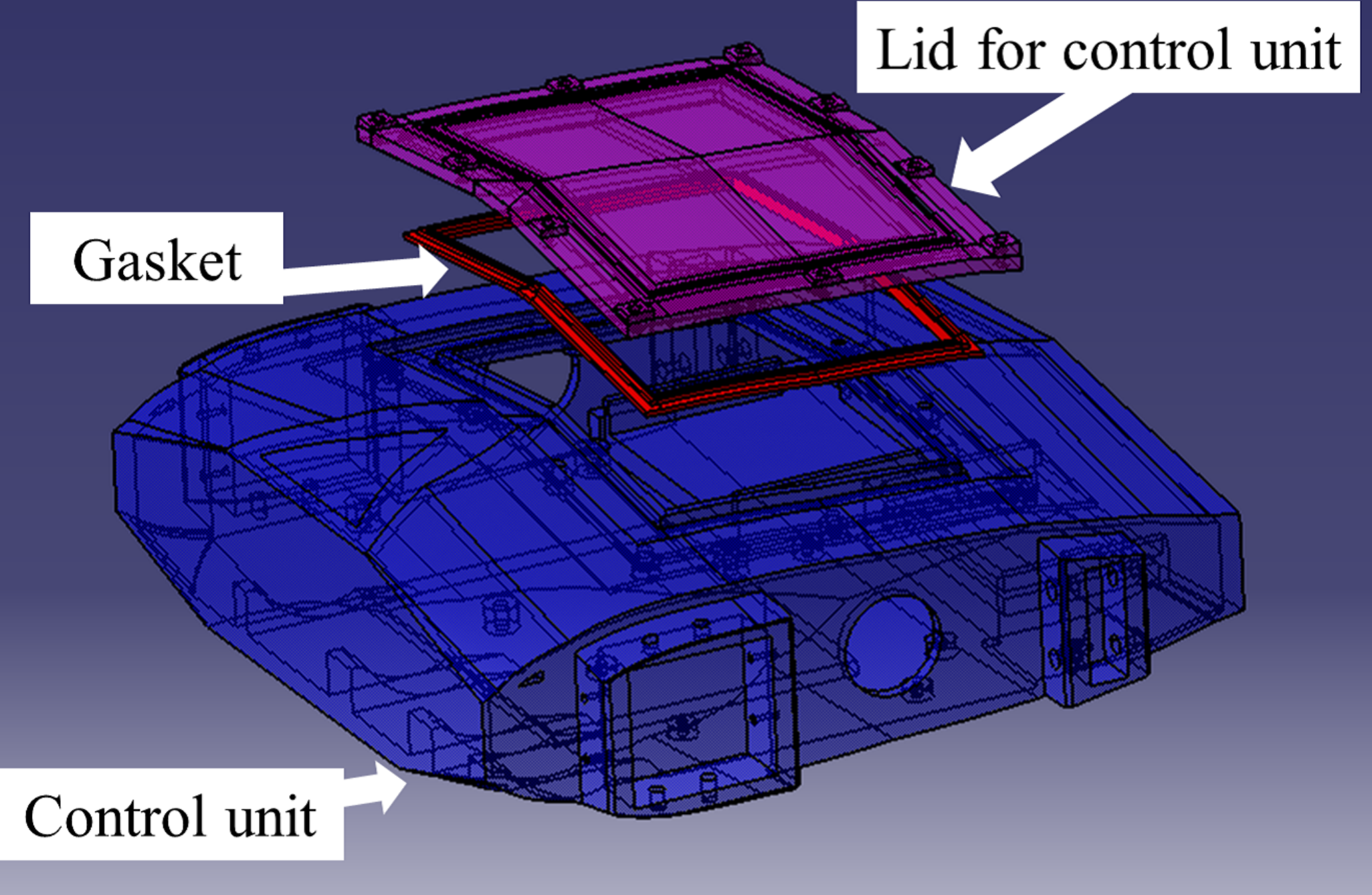}
      \subcaption{Design of control box}
      \label{fig:DesignControlBox}
    \end{minipage}
    \begin{minipage}[h]{\hsize}
      \vspace{4pt}
      \centering
      \includegraphics[height=0.45\hsize]{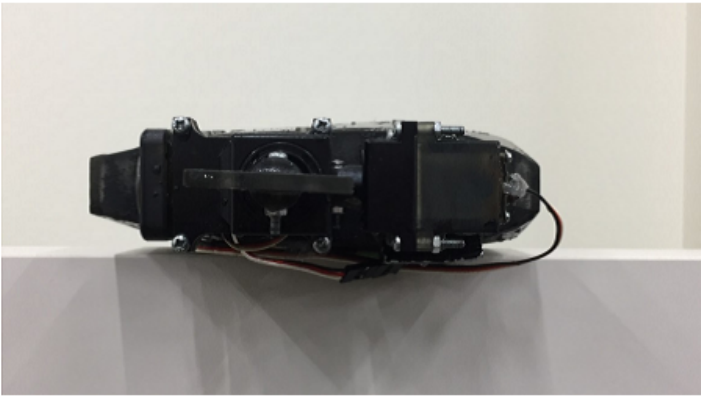}
      \subcaption{Fabricated of control box}
      \vspace{4pt}
      \label{fig:FabricatedControlBox}
   \end{minipage}
  \caption{Design and fabricated control box}
  \label{hogfe}
\end{figure}

\begin{figure}[h]
 \centering
  \includegraphics[height=0.8\hsize]{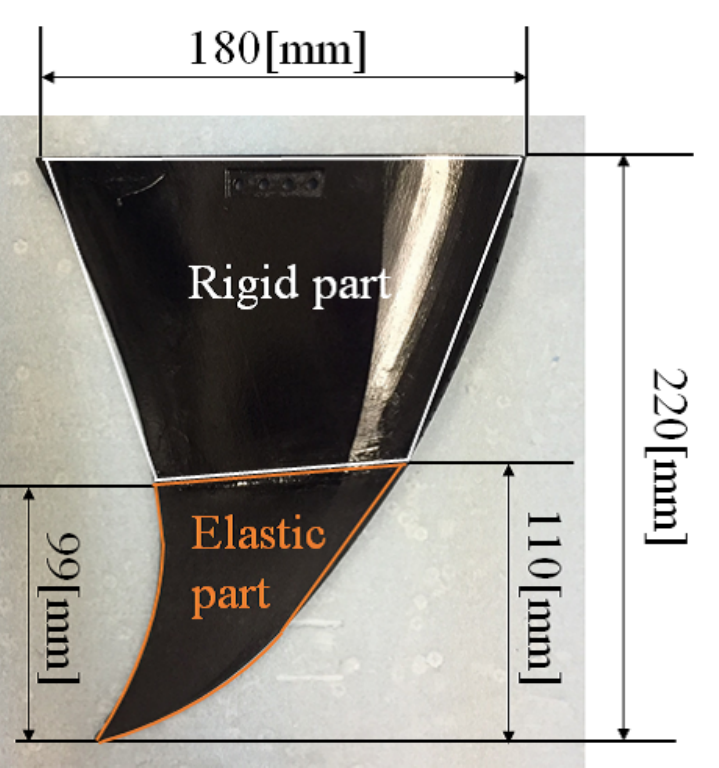}
  \caption{Developed pectoral fin}
  \label{fig:DevelopFin}
\end{figure}

We focus on the real manta-ray shape. According to Clark et al., the cross-sectional shape of the manta ray, when viewed from the side, forms a streamlined shape\cite{clark2006thrust}. 
This cross-sectional shape is similar to the NACA 0020 airfoil, which is one of the wing profiles used in airplanes. 
The advantage of this shape is that it reduces resistance from the fluid, allowing the manta ray to swim long distances with fewer flaps. By mimicking this shape, similar benefits are expected for the robot.
We designed and developed the robot's control box  shape to mimick the NACA 0020 airfoil as shown in {\bf{Fig.\ref{fig:DesignControlBox}}}. 
Pectoral fins are also designed based on the same airfoil ({\bf{Fig.\ref{fig:DevelopFin}}}).  The fins composed of a elastic part and a rigid part. 
The length of the elastic part is attached is 110 mm at the anterior edge from the tip of the fin and 99 mm  at the posterior edge. 
The size of the fin is 220 × 180 × 20 (length × width × maximum height) mm, and its weight is 117 g. The fins were fabricated using an Objet Eden260VS 3D printer (Stratasys). 
The rigid part of the fin was fabricated using RGD720 (acrylic resin, Shore hardness: 83-86 Hs), and the elastic part using RGD720 and TangoBlackPlus (rubber-like resin, Shore hardness: 26-28 Hs).

The Control box consists of a Raspberry Pi 3B, MPU9050, Arduino Nano, and a battery. 
The MPU9050 is connected to the Arduino Nano, which sends 3-axis gyro, acceleration, and magnetic data via serial communication.
Pressure sensor (LPS33HW) is mounted at sensor part. 
The LPS33HW is waterproof and capable of measuring atmospheric pressure, water pressure, and temperature. 
The sensor's measurement pressure range is 260 to 1,260 hPa, allowing for the measurement of water depths from approximately 0 to 260 cm at an atmospheric pressure of 1,000 hPa. 
The sensor's measurement accuracy is ±0.1 hPa (at a temperature of 25°C and a pressure range of 800 to 1,100 hPa). 

\subsection{System configuration}

\begin{figure}[h]
 \centering
  \includegraphics[height=0.8\hsize]{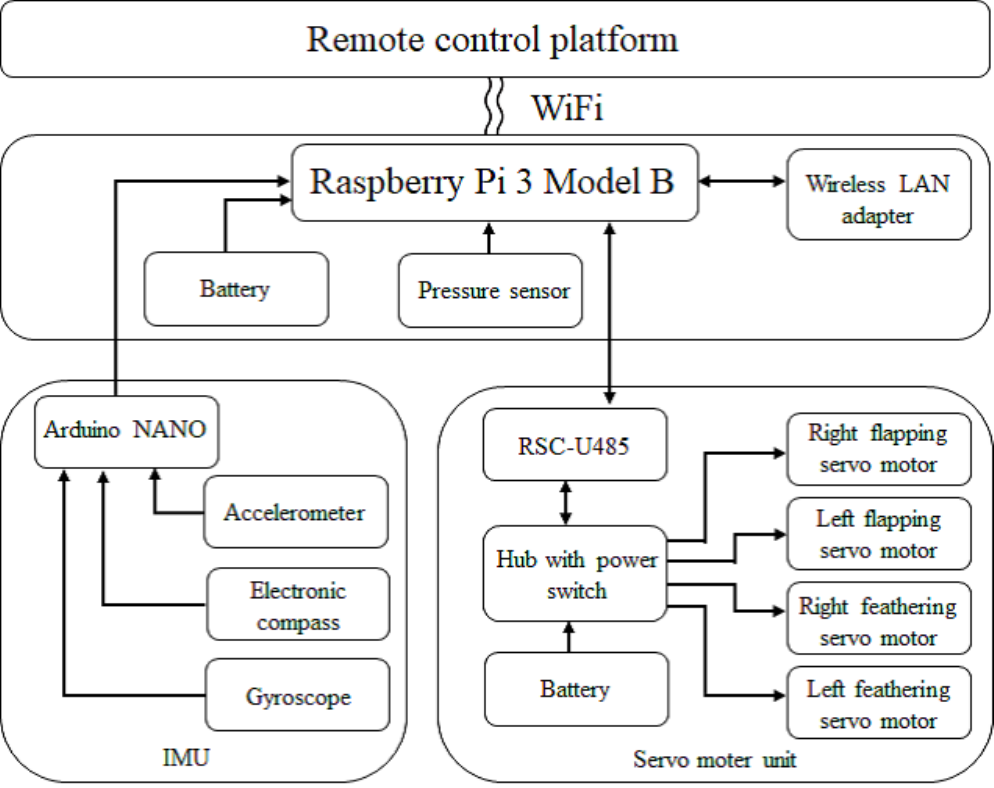}
  \caption{Control system diaglam}
  \label{fig:ContSys}
\end{figure}

The structure of the control system is shown in {\bf{Fig.\ref{fig:ContSys}}}. The Raspberry Pi 3 Model B is connected via SSH and communicates through Wi-Fi. Communication between the Raspberry Pi 3 Model B and the Arduino Nano is conducted via USB, and the pressure sensor is connected to the GPIO pins of the Raspberry Pi 3 Model B using jumper wires. Due to the weak signal strength of the built-in Wi-Fi in the Raspberry Pi 3 Model B, a USB Wi-Fi dongle (LAN-W300N/U2S by Logitec) is attached to the USB port of the Raspberry Pi 3 Model B. The angles of the servo motors in the left and right pectoral fins are determined based on the measurements obtained from the IMU sensor and the pressure sensor. The batteries are installed separately for the Raspberry Pi 3 Model B and the servo motors, and provide approximately 50 minutes of operation time for the robot.

\section{Experiment setup}

\begin{figure}[h]
 \centering
  \includegraphics[height=0.7\hsize]{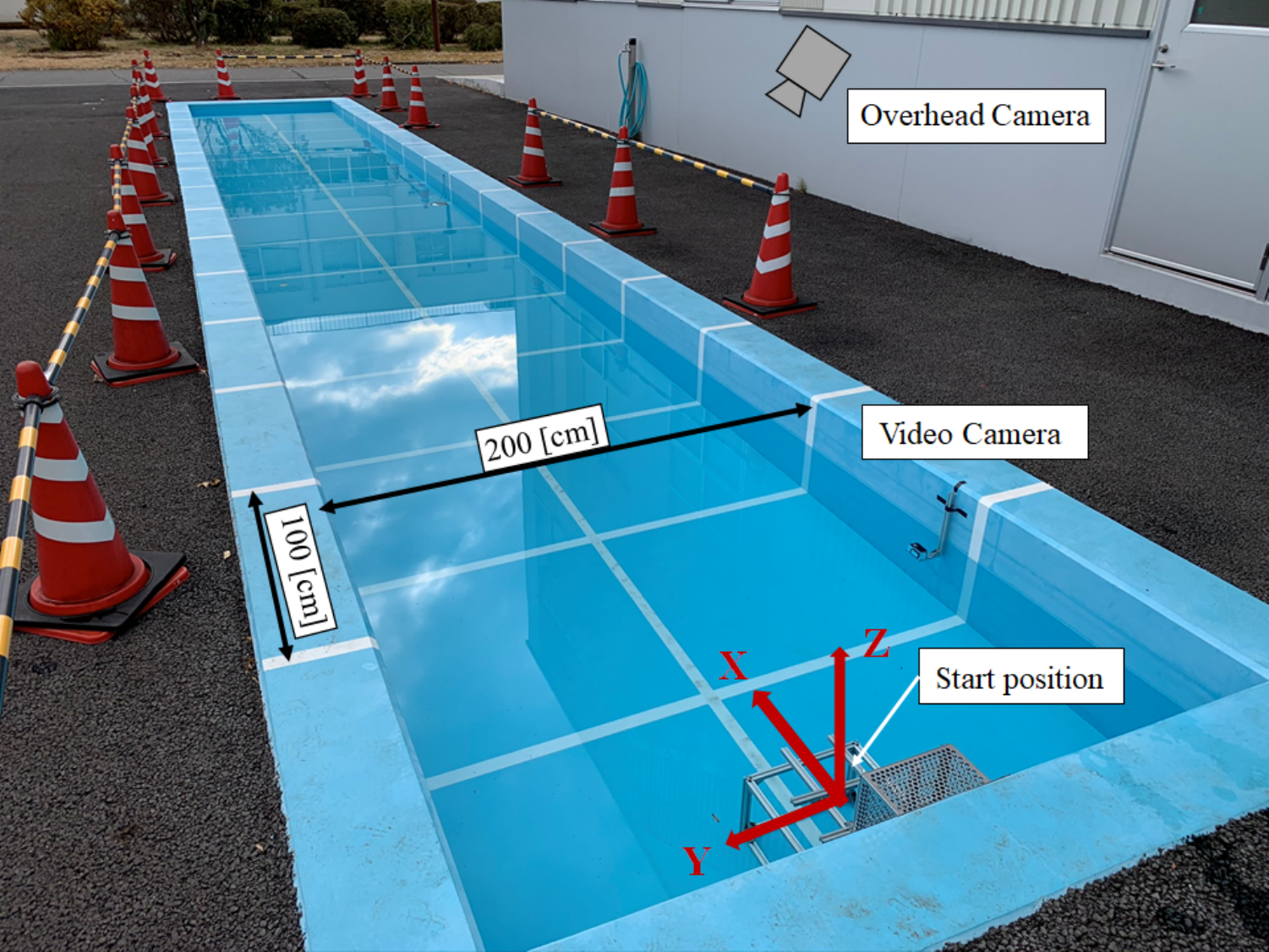}
  \caption{Experiment setting}
  \label{fig:Pool}
\end{figure}

We conducted two experiments. 
The first experiment focused on the swimming motion, where the robot moved in a straight trajectory. 
The second experiment investigated the diving motion. Both experiments were conducted in the pool shown in {\bf{Fig.\ref{fig:Pool}}}. 
The dimensions of the pool are 1500 cm in length, 200 cm in width, and 48.5 cm in depth.
The robot was placed on a starting platform positioned within the pool, from which all experiments commenced.
We positioned video cameras above the pool and near the water's surface to capture the movements of the robot. 

We defined the robot's motion (flapping angles $\theta_{fl}$ and feathering angles $\theta_{fe}$) through preliminary experiments.
$\theta_{fe}$ and $\theta_{fl}$ are defined as equation (\ref{eq:fl}) and (\ref{eq:fe}). 

\begin{align}
    \theta_{fl} &= \theta_{\rm{flmax}} \sin(2\pi tf) \label{eq:fl}\\
    \theta_{fe} &= \theta_{\rm{femax}} \sin\left(2\pi tf - \frac{\pi}{2}\right) \label{eq:fe}
\end{align}

Here, $\theta_{\rm{fl}}$, $\theta_{\rm{fe}}$, $\theta_{\rm{flmax}}$, $\theta_{\rm{femax}}$, $f$, and $t$ represent the flapping angle [deg], feathering angle [deg], maximum flapping angle [deg], maximum feathering angle [deg], frequency [Hz], and time [s], respectively. The phase difference between the flapping and feathering motions was set to 90 degrees, which generates the maximum thrust.


\section{Swimming motion experiment}

\subsection{Swimming motion on the water surface}

The maximum flapping angles, ($\theta_{\text{fmax}}$), and feathering angles, ($\theta_{\text{femax}}$), was selected, as shown in {\bf{Table \ref{tb:parathata}}}.
The frequency of the pectoral fin motion was determined through trial and error.
The developed robot struggles to move straight without control due to slight deviations in the timing of the actuation on either side and the placement of its center of gravity. Therefore, using angular data from the IMU, straight motion is achieved through PD control. 
The phase difference between flapping and feathering movements in the motions is always 90[deg]. 
Experiments were conducted three times each with and without PD control for the parameters of the flapping movements shown in {\bf{Table \ref{tb:parathata}}}. 
The differences between the calculated turning angles and the target angles, as well as the formulas for the amplitudes of the left and right flapping movements, are shown in Equations (\ref{eq:3}), (\ref{eq:4}), and (\ref{eq:5}), respectively.
The control equations for the robot's fins are as follows:

\begin{align}
    \theta_{\text{err}} &= \theta_{\text{yaw}} - \theta_{\text{d}} \label{eq:3} \\
    \theta_{\text{flR}} &= \theta_{\text{flR0}} - k_{\text{p}} \theta_{\text{err}} - k_{\text{d}} (\theta_{\text{err}} - \theta_{\text{err-1}}) \label{eq:4} \\
    \theta_{\text{flL}} &= \theta_{\text{flL0}} + k_{\text{p}} \theta_{\text{err}} + k_{\text{d}} (\theta_{\text{err}} - \theta_{\text{err-1}}) \label{eq:5}
\end{align}

Here, \(\theta_{\text{err}}\) [deg] is the error from the target angle, \(\theta_{\text{yaw}}\) [deg] is the angle the robot has rotated around the yaw axis from its initial posture, and \(\theta_{\text{d}}\) [deg] is the robot's target angle. \(\theta_{\text{flR0}}\) [deg] and \(\theta_{\text{flL0}}\) [deg] are the initial flapping angles for the right and left fins, respectively, when the error \(\theta_{\text{err}} = 0\). The terms \(k_p\) and \(k_d\) represent the proportional and derivative gains, respectively.

\begin{table}[b]
    \centering
    \caption{Parameters of swimming motion on the water surface}
    \label{tb:parathata}
    \scalebox{0.9}{
        \begin{tabular}{|c|c|c|c|}
        \hline
        Number &Flapping angle [deg] & Feathering angle [deg] & Frequency [Hz] \\ \hline
        1 & 30 & 45 & 0.75 \\ \hline
    \end{tabular}
    }
\end{table}

\begin{figure}[h]
 \centering
  \includegraphics[height=0.6\hsize]{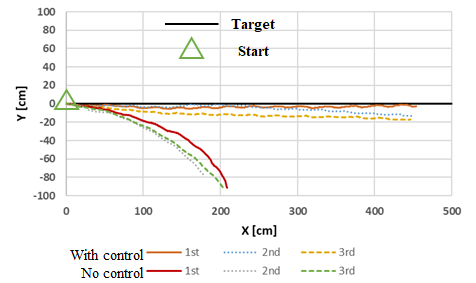}
  \caption{Robot trajectories with and without PD control on the water surface}
  \label{fig:PDStraight}
\end{figure}

{\bf{Figure\ref{fig:PDStraight}}} shows the results of a straight-line motion experiment using the parameter for the underwater robot. 
The target trajectory is indicated by the black line. The experiment was conducted three times, and the results are presented in the graph with PD control ("with control") and without PD control ("No control"). Additionally, {\bf{Table \ref{tb:PDStraight}}} displays the maximum error, average error, and standard deviation between the target trajectory and the actual trajectory when PD control was employed. The results confirm that integrating PD control improves the straight-line motion of the robot.

\begin{figure}[h]
 \centering
  \includegraphics[height=0.6\hsize]{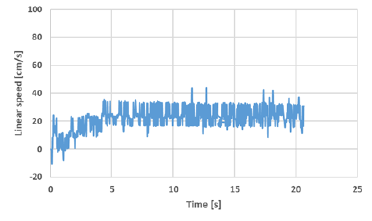}
  \caption{Linear speed of swimming motion}
  \label{fig:lin_speed}
\end{figure}

\begin{figure}[h]
 \centering
  \includegraphics[height=0.6\hsize]{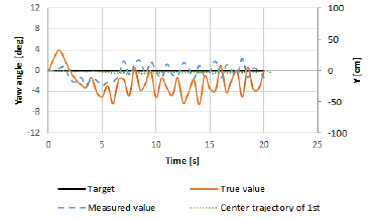}
  \caption{Yaw angle transition in 1st trial with PD control}
  \label{fig:yawangle}
\end{figure}

{\bf{{Figure~\ref{fig:lin_speed}}}} shows the swimming velocity in Experiment 1 with PD control. After 5 seconds, the velocity stabilizes. The average speed was approximately 20 cm/s, allowing the robot to swim a distance of 500 cm in about 20 seconds.
Additionally, as shown in {\bf{Fig.~\ref{fig:yawangle}}}, the variation in the yaw angle measured by the IMU and the yaw angle measured by the camera (True value) are presented. Due to the PD control, even looking at the true values, the variation in yaw angle is contained within approximately -6 to +4 deg. Although the angle calculated by the IMU differs from the true value, it captures the variation in angle, contributing to the effectiveness of the PD control.

\begin{table}[h]
    \centering
    \caption{Error Metrics and Standard Deviation}
    \label{tb:PDStraight}
    \begin{tabular}{|c|c|c|c|}
        \hline
        & Trial 1 & Trial 2 & Trial 3 \\ \hline
        Mean Error [cm] & 2.8 & 4.5 & 10.1 \\ \hline
        Maximum Error [cm] & 5.2 & 13.2 & 17.5 \\ \hline
        Standard Deviation [cm] & 1.2 & 3.7 & 5.0 \\ \hline
    \end{tabular}
\end{table}

\subsection{Swimming motion under the water}

Using the parameters listed in \textbf{Table~\ref{tb:parathata2}}, the robot was instructed to perform a straight-line motion while executing a diving motion. In preliminary experiments, the flapping motion with Parameter in \textbf{Table~\ref{tb:parathata}}, which involves both flapping and feathering angles at (45deg), generated the highest propulsion speed. The flapping motion with Parameter Number 2, with a flapping angle of 30deg and a feathering angle of 45deg, produced a lower propulsion speed than the first, but had smaller variations between maximum and minimum speeds, indicating better energy efficiency.

\begin{figure}[h]
 \centering
  \includegraphics[height=0.6\hsize]{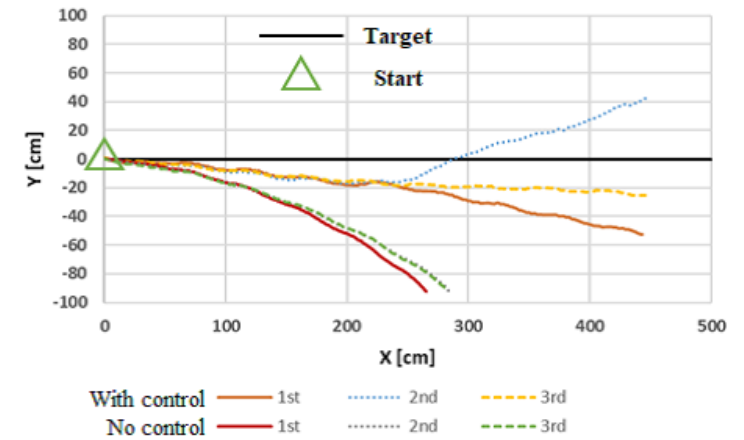}
  \caption{Robot trajectories with and without PD control during diving motion}
  \label{fig:PDStraight_diving}
\end{figure}

\textbf{Figure~\ref{fig:PDStraight_diving}} shows the results of the straight-line motion while diving, comparing the scenarios with and without PD control. At 200 cm, the robot collided with the bottom of the tank, significantly altering its trajectory. This incident highlights the need for additional control methods beyond PD control to handle significant disturbances. However, up to 200 cm, a certain degree of straight-line motion was maintained, confirming the effectiveness of PD control to some extent.

\begin{figure}[h]
 \centering
  \includegraphics[height=0.6\hsize]{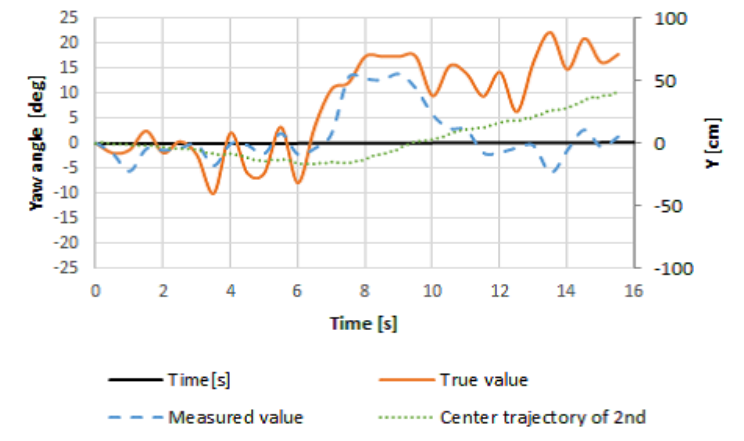}
  \caption{Yaw angle transition in 2nd trial with PD control}
  \label{fig:Yawangle_div}
\end{figure}

\textbf{Figure~\ref{fig:Yawangle_div}} shows the changes in yaw angle during a straight-line motion while diving. At approximately 6 seconds, the robot collided with the bottom of the tank, resulting in a significant change in yaw angle. The values changed at this time, leading to discrepancies between the true values and those measured by the IMU, which could not be controlled.

\begin{table}[h]
    \centering
    \caption{Parameters of swimming motion with diving}
    \label{tb:parathata2}
    \scalebox{0.9}{
        \begin{tabular}{|c|c|c|c|}
        \hline
        Number &Flapping angle [deg] & Feathering angle [deg] & Frequency [Hz] \\ \hline
        2 &30 & 45 & 0.75 \\ \hline
    \end{tabular}
    }
\end{table}

\begin{table}[ht]
\centering
\caption{Trajectories error of experiment in \textbf{Fig.~\ref{fig:PDStraight_diving}}}
\label{tb:exp_error}
    \begin{tabular}{|c|c|c|c|}
        \hline
        \textbf{Measure} & \textbf{First} & \textbf{Second} & \textbf{Third} \\ \hline
        Mean Error [cm] & 19.4 & 2.7 & 13.4 \\ \hline
        Maximum Error [cm] & 52.4 & 42.6 & 25.4 \\ \hline
        Standard Deviation [cm] & 15.3 & 16.4 & 7.7 \\ \hline
    \end{tabular}
\end{table}

\section{Experiments of diving motion}

This experiment was conducted in the same pool as described in the previous chapter. The robot was placed on a starting platform and initially performed a straight swimming motion. 
After a set period, the robot began a diving motion while continuing to swim straight. 
The swimming process was recorded using two video cameras, one positioned above the pool and another inside the pool. 
After the experiment, the robot's trajectory was calculated from the recorded video footage. 
The experiment concluded when the robot collided with the pool's wall or passed a position 500 cm from the edge of the pool. 
The parameters used in the experiment were flapping and feathering angles of 30 and 45 deg, respectively, a phase difference of 90 deg, a frequency of 0.75 Hz, and a target depth of 10 cm, with each trial conducted three times.

\begin{figure}[h]
 \centering
  \includegraphics[height=0.6\hsize]{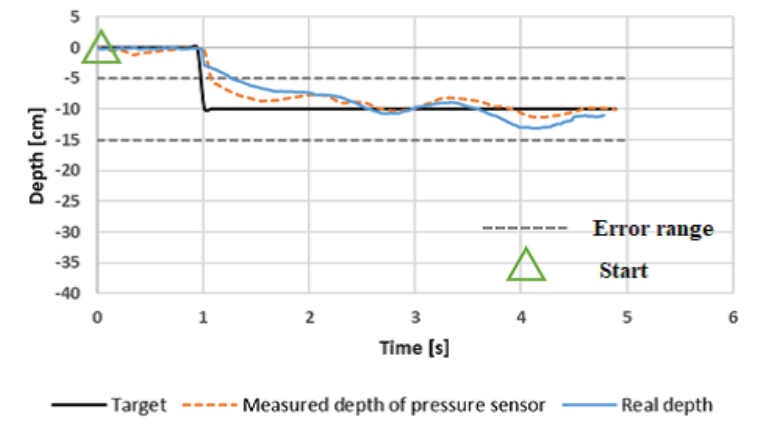}
  \caption{Yaw angle transition in 2nd trial with PD control}
  \label{fig:depth_cont}
\end{figure}

\begin{figure}[h]
 \centering
  \includegraphics[height=0.6\hsize]{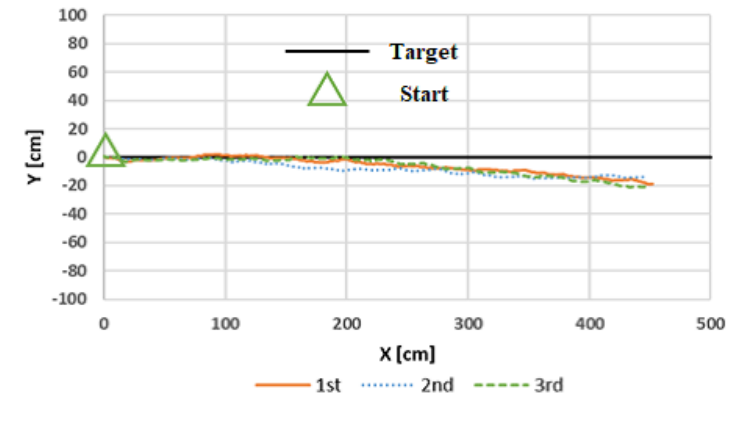}
  \caption{Yaw angle transition in 2nd trial with PD control}
  \label{fig:straight_divcontrol}
\end{figure}

\begin{figure}[h]
 \centering
  \includegraphics[height=0.5\hsize]{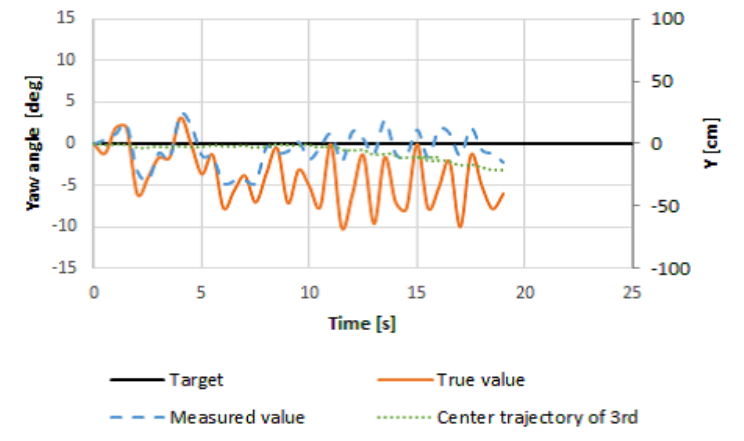}
  \caption{Yaw angle transition in 3rd trial with depth control}
  \label{fig:Yawdepth_cont}
\end{figure}

The trajectories of the robot's straight and diving motions obtained from the experiment are shown in {\bf{Fig.~\ref{fig:depth_cont}}} and {\bf{Fig.~\ref{fig:straight_divcontrol}}}, respectively. Notably, {\bf{Fig.~\ref{fig:depth_cont}}} shows the results of the third trial. The average swimming speed during straight motion was 22 cm/s, consistent with the results in Section 4, indicating that if the flapping motion parameters remain the same, the speed does not change even during diving motion. From {\bf{Fig.~\ref{fig:straight_divcontrol}}}, it is evident that in all three experiments, the robot deviated to the right with respect to the negative Y-axis direction, i.e., the forward direction. As shown in {\bf{Table~\ref{tab:error_statistics}}}, the maximum error was 21.4 cm. This error can be attributed to the pitch motion performed during the diving swim, leading to significant cumulative errors in the integration of the angular velocity by the IMU. {\bf{Fig.~\ref{fig:Yawdepth_cont}}} shows a representative result of the yaw angle measured by the IMU compared with the yaw angle determined from the video footage, specifically the third trial graph.

\begin{table}[h]
\centering
\caption{Trajectories error of diving control}
    \begin{tabular}{|c|c|c|c|}
        \hline
         & First & Second & Third \\
        \hline
        Mean error [cm] & 5.5 & 7.1 & 5.8 \\
        \hline
        Maximum error [cm] & 19.0 & 15.5 & 21.4 \\
        \hline
        Standard deviation [cm] & 5.6 & 5.1 & 6.3 \\
        \hline
    \end{tabular}
\label{tab:error_statistics}
\end{table}

\begin{table}[h]
    \centering
    \caption{Yaw angle error of diving control}
    \begin{tabular}{|c|c|c|c|}
    \hline
    Mean error [cm] & 3.1 & 2.1 & 3.3 \\
    \hline
    Maximum error [cm] & 12.8 & 7.5 & 8.6 \\
    \hline
    Standard deviation [cm] & 3.1 & 1.9 & 2.6 \\
    \hline
    \end{tabular}
    \label{tab:error_statistics_new}
\end{table}

\section{Conclusion}

The development and experimental verification of the biomimetic manta ray robot demonstrate its capability for underwater autonomous exploration and ecological monitoring. By employing flapping motion for propulsion, the robot minimizes seabed disturbance and improves efficiency compared to traditional screw propulsion systems. Experimental results indicate that the robot can achieve stable swimming and diving motions with PD control. We believe thar the design is particularly suited for applications in environments like aquariums and fish nurseries. This study highlights the practical advantages of bio-inspired robotic designs and suggests its potential for broadly applications in various underwater settings. Future work will aim to improve the robot's maneuverability and operational capabilities in more complex and dynamic underwater environments.

\section*{Acknowledgment}
This work was supported by the Sasakawa Scientific Research Grant from The Japan Science Society.
\footnotesize

\bibliographystyle{unsrt} 
\bibliography{ref} 

@inproceedings{Habib2013EngineeringCD,
  title={Engineering Creative Design in Robotics and Mechatronics},
  author={Maki K. Habib and J{\~o}ao Paulo Davim and Lindsay Johnston and Joel Gamon and Jennifer Yoder and Adrienne Freeland and Christine Smith and Kayla Wolfe and Christina Henning and Jason Mull},
  year={2013},
  url={https://api.semanticscholar.org/CorpusID:6545581}
}

@article{salazar2018classification,
  title={Classification of biological and bioinspired aquatic systems: A review},
  author={Salazar, R and Fuentes, V and Abdelkefi, A},
  journal={Ocean Engineering},
  volume={148},
  pages={75--114},
  year={2018},
  publisher={Elsevier}
}

@article{bhlpart203769,
	title = {The locomotion of fishes},
	journal = {Zoologica : scientific contributions of the New York Zoological Society},
	volume = {4},
	copyright = {In Copyright. Digitized with the permission of the rights holder},
	url = {https://www.biodiversitylibrary.org/part/203769},
	publisher = {New York, },
	author = {Breder and  Charles M. (Charles Marcus)},
	year = {1926-09-29},
	pages = {159--297},
}

@incollection{LINDSEY19781,
title = {1 - Form, Function, and Locomotory Habits in Fish},
editor = {W.S. Hoar and D.J. Randall},
series = {Fish Physiology},
publisher = {Academic Press},
volume = {7},
pages = {1-100},
year = {1978},
booktitle = {Locomotion},
issn = {1546-5098},
doi = {https://doi.org/10.1016/S1546-5098(08)60163-6},
url = {https://www.sciencedirect.com/science/article/pii/S1546509808601636},
author = {C.C. Lindsey}
}

@article{claudio2020design,
  title={Design of a bio-inspired manta ray robot},
  author={Claudio, Ivan},
  year={2020}
}

@article{wang2024soft,
  title={Soft Manta Ray Robot Based on Bilateral Bionic Muscle Actuator},
  author={Wang, Ruiqian and Zhang, Chuang and Zhang, Yiwei and Yang, Lianchao and Qin, Hengshen and Zhang, Qi and Yang, Yongliang and Liu, Lianqing},
  journal={IEEE Robotics and Automation Letters},
  year={2024},
  publisher={IEEE}
}

@article{zhou2010better,
  title={Better endurance and load capacity: an improved design of manta ray robot (RoMan-II)},
  author={Zhou, Chunlin and Low, Kin-Huat},
  journal={Journal of Bionic Engineering},
  volume={7},
  pages={S137--S144},
  year={2010},
  publisher={Elsevier}
}

@article{asada2024development,
  title={Development of a Manta Ray Robot with Underwater Walking Function},
  author={Asada, Takumi and Furuhashi, Hideo},
  journal={Ocean Engineering},
  volume={308},
  pages={118261},
  year={2024},
  publisher={Elsevier}
}

@inproceedings{osorio2023manta,
  title={Manta Ray inspired multistable soft robot},
  author={Osorio, Juan C and Tinsley, Chelsea and Tinsley, Kendal and Arrieta, Andres F},
  booktitle={2023 IEEE International Conference on Soft Robotics (RoboSoft)},
  pages={1--6},
  year={2023},
  organization={IEEE}
}

@article{cao2023realization,
  title={Realization and Online Optimization for Gliding and Flapping Propulsion of a Manta Ray Robot},
  author={Cao, Yonghui and Cao, Yingzhuo and Ma, Shumin and Li, Xinhao and Qu, Yilin and Cao, Yong},
  journal={Journal of Marine Science and Engineering},
  volume={11},
  number={11},
  pages={2173},
  year={2023},
  publisher={MDPI}
}

@article{liu2023design,
  title={Design and realization of a novel robotic manta ray for sea cucumber recognition, location, and approach},
  author={Liu, Yang and Liu, Zhenna and Yang, Heming and Liu, Zheng and Liu, Jincun},
  journal={Biomimetics},
  volume={8},
  number={4},
  pages={345},
  year={2023},
  publisher={MDPI}
}

@article{clark2006thrust,
  title={Thrust production and wake structure of a batoid-inspired oscillating fin},
  author={Clark, Richard P and Smits, Alexander J},
  journal={Journal of fluid mechanics},
  volume={562},
  pages={415--429},
  year={2006},
  publisher={Cambridge University Press}
}
\normalsize
\end{document}